\documentclass{article}

\usepackage{microtype}
\usepackage{graphicx}
\usepackage{subfigure}
\usepackage{booktabs} 

\usepackage{hyperref}

\usepackage{amsmath}
\usepackage{amssymb}
\usepackage{makecell}
\usepackage{multirow}

\usepackage{icml2019}

\icmltitlerunning{A Simple Method to Reduce Off-chip Memory Accesses on Convolutional Neural Networks}

\begin{document}

\twocolumn[
\icmltitle{A Simple Method to Reduce Off-chip Memory Accesses \\ on Convolutional Neural Networks}



\icmlsetsymbol{equal}{*}

\begin{icmlauthorlist}
\icmlauthor{Doyun Kim}{equal,S.LSI}
\icmlauthor{Kyoung-Young Kim}{equal,S.LSI}
\icmlauthor{Sangsoo Ko}{S.LSI}
\icmlauthor{Sanghyuck Ha}{S.LSI}
\end{icmlauthorlist}

\icmlaffiliation{S.LSI}{System LSI Division, Samsung Electronics, Korea}
\icmlcorrespondingauthor{Doyun Kim}{dyun.kim@samsung.com}

\icmlkeywords{Simple Method, NPU, efficiency, SRAM, DRAM, off-chip memory, on-chip memory, Deep Learning, ICML}

\vskip 0.3in
]



\printAffiliationsAndNotice{\icmlEqualContribution} 

\begin{abstract}
For convolutional neural networks, a simple algorithm to reduce off-chip memory accesses is proposed by maximally utilizing on-chip memory in a neural process unit. Especially, the algorithm provides an effective way to process a module which consists of multiple branches and a merge layer. For Inception-V3 on Samsung's NPU in Exynos, our evaluation shows that the proposed algorithm makes off-chip memory accesses reduced by 1/50, and accordingly achieves 97.59\% reduction in the amount of feature-map data to be transferred from/to off-chip memory.
\end{abstract}

\section{Introduction}
\label{sec:itroduction}

Recent achievements in image processing tasks such as image recognition, object detection, and scene segmentation have been coupled with the application of deep convolutional networks \cite{szegedy2015going, ren2015faster, long2015fully}. As the need for more complex networks increases, we get faced with several implementation issues, i.e. real time processing, limited power budget, and memory bandwidth. For the issues to get resolved, various approaches have been investigated in both cloud and mobile applications; low-precision \cite{courbariaux2014training,courbariaux2015binaryconnect,hubara2016quantized,gupta2015deep,gysel2016hardware,judd2015reduced,lin2016fixed, kim2018convolutional}, network compression \cite{han2015deep}, \cite{han2016dsd}, and small network design \cite{iandola2016squeezenet}, \cite{howard2017mobilenets}.

Another remarkable trend is to execute deep convolutional networks on mobile platforms, and it is getting important by concerns about response time, dependency on an internet connection, privacy, and security. Many companies and research groups have been recently developing notable hardware accelerators called as a neural processing unit (NPU) \cite{song2018makalu, zhang2016cambricon, Marvin2018Online}. They tried to develop energy-efficient NPUs based on novel algorithms such as exploiting network sparsity for high utilization of multiply/accumulate (MAC) units or quantizing networks to reduce the power of MAC units. To operate CNNs on a NPU, it is needed to access memory hugely to read and write weight and feature-map data. In \cite{han2016eie}, it was shown that the total energy is dominated by the required memory access if there is no data reuse, and that the energy cost of on-chip memory (SRAM in \cite{han2016eie}) is 128 times better than one of off-chip memory (DRAM in \cite{han2016eie}). Since NPUs for a mobile platform have the limited amount of on-chip memory, however, it is not easy to maintain NPU efficiency for most applications. Therefore, we reasoned that reducing off-chip memory accesses by utilizing on-chip memory maximally can be one of the most powerful solutions to increase the efficiency of NPUs.

\begin{figure}
\begin{center}
\includegraphics[width=1.0\linewidth]{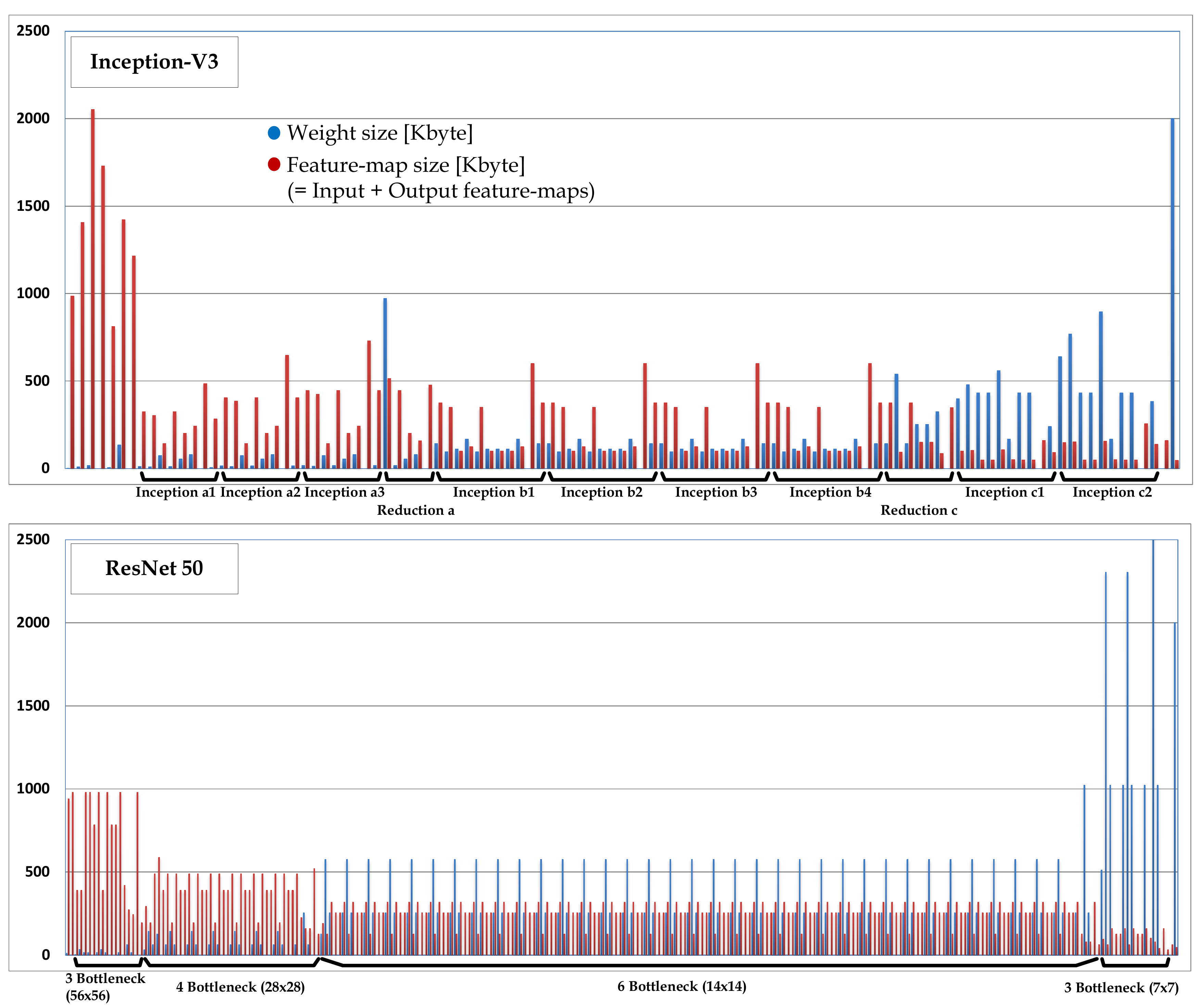}
\end{center}
   \caption{Sizes of weights and feature-maps according to indexes of layers for Inception-V3 and ResNet-50.}
\label{fig:001}
\end{figure}

To achieve the high utilization of on-chip memory, we needed to know what series of operations are required in NPUs for processing the convolution layer and to choose the most efficient one among possible series of operations. We focused on a series of operations which fetches weights in minimal increments and executes convolution with the weights for the feature-maps of all input channels. Such a series of operations makes the realization of convolution simple by leaving out the consideration about how to locate weights in on-chip memory.
Figure \ref{fig:001} shows the required memory size of weights and feature-maps according to the index of layer, for Inception-V3 and ResNet-50. The size of weights increases as the layer index increases, whereas the size of feature-maps decreases as the layer index increases. From the patterns of sizes, we thought that it would be beneficial to locate feature-maps fully in on-chip memory and to manage them efficiently if the size of feature-maps is small enough to be located in on-chip memory. In recent architectures of convolutional neural networks, moreover, a concept of module or block has been introduced to computer vision applications for higher representational power of neural networks. The important thing of such observation is that modules or blocks are repeatedly used in a network just after feature-maps are scaled down for reduction of computational cost, as shown in Figure \ref{fig:001}. Depending on the size of feature-maps and the characteristics of modules, therefore, we were able to find a simple way to minimize off-chip memory accesses by efficiently managing on-chip memory for modules.

In this paper, we propose a simple method to support energy-efficient and real-time processing of NPUs through the reduction of off-chip memory accesses.
Firstly, the algorithm detects certain types of modules or blocks by graph interpretation of the whole network. Then, several regions of on-chip memory are assigned to reuse feature-maps during a module or a block processing. In order to utilize on-chip memory maximally, moreover, we also propose a branch-reordering algorithm and two branch-processing algorithms. By combining the proposed algorithms, we can effectively cut down off-chip memory accesses for convolutional neural networks such as Inception-V3 and ResNet. The rest of the paper is organized as follows. Section \ref{sec:definition} describes the module we define. Then, in Section \ref{sec:algorithm}, we introduce the proposed algorithms to reduce off-chip memory accesses. Section \ref{sec:evaluation} shows evaluation results for a representative network, Inception-V3 \cite{szegedy2015going, szegedy2016rethinking}. Finally, Section \ref{sec:conclusion} makes a conclusion.

\section{Definition of Module}
\label{sec:definition}

After Network-in-Network was proposed in \cite{lin2013network} in order to increase the representational power of neural networks, a concept of module or block is getting popular in convolution neural networks for computer vision applications like Inception network \cite{szegedy2015going, szegedy2016rethinking}, ResNet \cite{he2016deep}, MobileNet V2 \cite{sandler2018mobilenetv2}, SqueezeNet \cite{iandola2016squeezenet}, ShuffleNet \cite{ma2018shufflenet}, and MnasNet \cite{tan2018mnasnet}. In this section, we define module and specify which modules can be utilized in the algorithms we will propose among the modules defined for various networks.

In general, a module used in convolutional neural networks can be one of the directed acyclic graphs (DAGs), which is a finite directed graph without directed cycles. That is, it consists of finite multiple layers and edges, with each edge directed from one layer to another, such that there is no way to start at any layer $\bold{A}$ and follow a consistently-directed sequence of edges that eventually loops back to $\bold{A}$ again. For a convolutional neural network including many modules, moreover, it can be viewed as a large DAG with multiple DAGs.

In the paper, we consider only the limited structure of DAGs satisfying the following conditions: 1) the module has multiple branches within itself, 2) the module has to include at least a merge layer, and 3) the type of merge layers should be either concatenation or element-wise summation. In addition, 4) we do not cover a large-sized module configured by a long skip-connection or having lots of layers inside, which has been frequently used in neural networks to extract multi-scaled features. This is because there seems no efficient way to utilize on-chip memory for large-sized modules. When we focused on only the modules satisfying the four conditions mentioned above, we could reach a sub-optimal but simple solution to reduce off-chip memory accesses even if the proposed algorithms did not cover all kinds of neural networks.

\begin{figure}[]
\begin{center}
\includegraphics[width=1.0\linewidth]{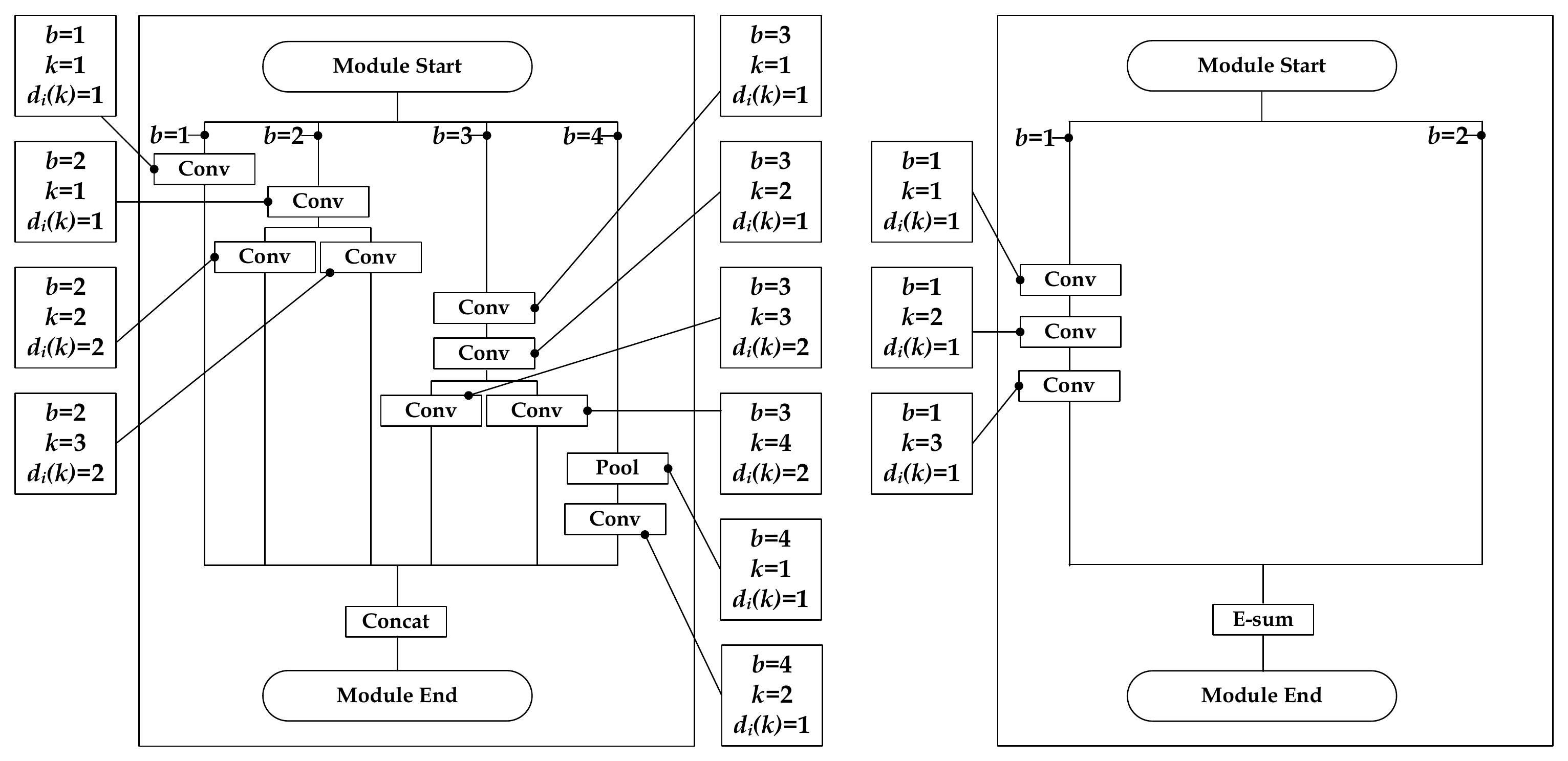}
\end{center}
   \caption{Illustration of various modules satisfying four conditions in Section \ref{sec:definition}. Here, $b$ and $k$ denote the indexes of a branch and a layer for the module, And $d_i(k)$ is the maximum depth of module at the $k^{th}$ layer.}
\label{fig:003}
\end{figure}

Figure \ref{fig:003} shows the examples of modules satisfying the required conditions and introduces some useful indices to explain the algorithms proposed in this paper, where $b$ denotes the index of branch in a module, $k$ is the index of layer in a branch, and $d_i(k)$ denotes the maximum depth of modules for the $k^{th}$ layer. In other words, it means $k^{th}$ layer is included in the $d_i(k) ^{th}$ module. With these parameters, we explain two types of modules, each having a different kind of merge layer as follows:

\textbf{Concatenation based module:} this type of module includes the $k$-th layer on the $b$-th branch, and a concatenation layer at the end of module, as shown in the left plot of Figure \ref{fig:003}. Sometimes the module also includes the sub-module which is a module in a module like as the Inception-C type. The representative neural networks having the concatenation based module are several versions of Inception Networks \cite{szegedy2015going, szegedy2016rethinking} and SqueezeNet \cite{iandola2016squeezenet}. In Figure \ref{fig:003}, $Conv$ and $Pool$ mean a convolutional layer and a pooling layer, respectively.

\textbf{Element-wise adder based module:} this type of module has two branches, multiple layers, and an element-wise adder at the end. The representative neural networks with the element-wise adder based module are ResNet \cite{he2016deep} and MobileNet V2 \cite{sandler2018mobilenetv2}.


\section{Algorithm for Effective Module Processing}
\label{sec:algorithm}

\begin{figure}[]
\begin{center}
\includegraphics[width=0.6\linewidth]{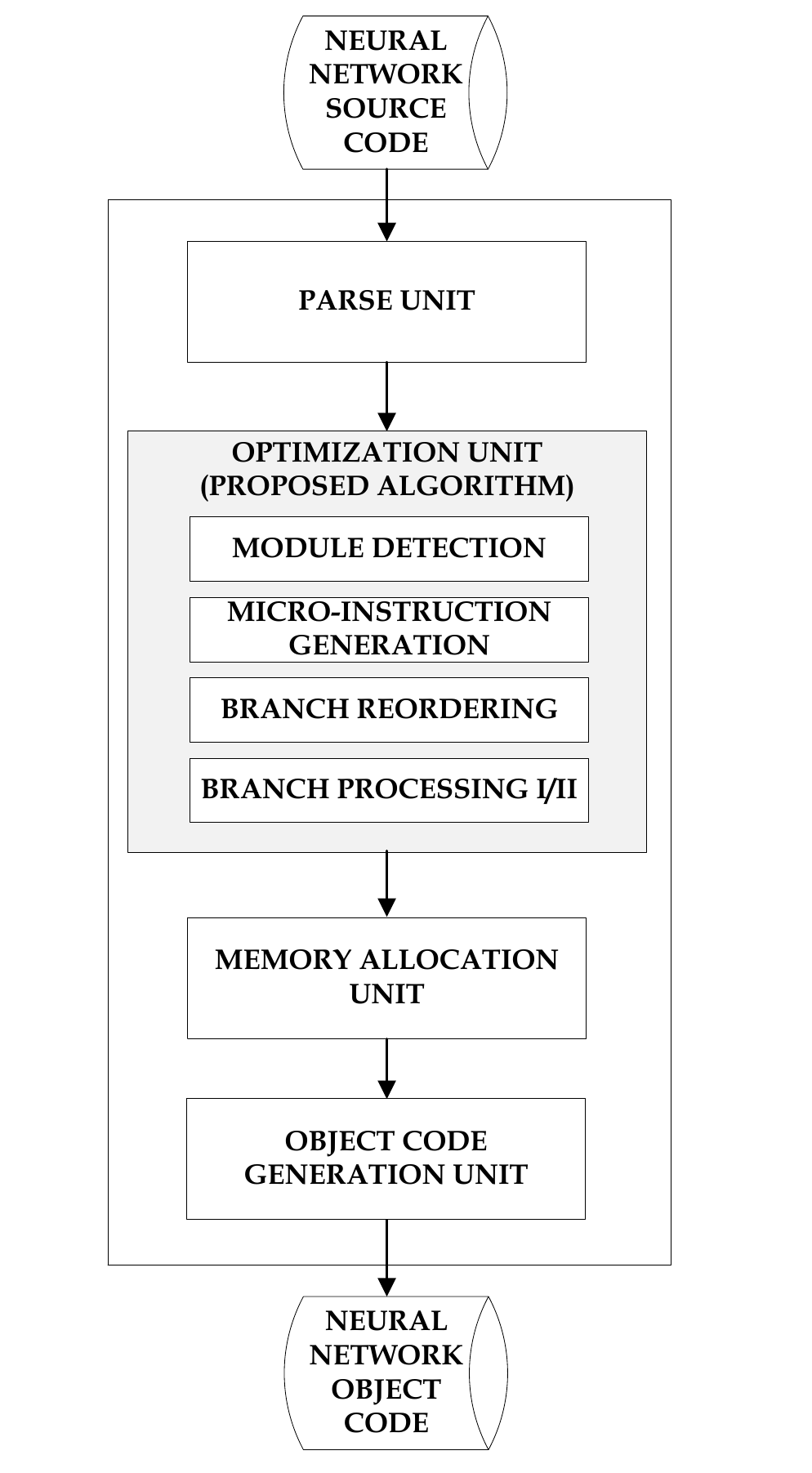}
\end{center}
   \caption{Overall flowchart of the compiler optimization including the proposed algorithm.}
\label{fig:004}
\end{figure}

The flowchart demonstrated in Figure \ref{fig:004} shows an overall process of compiler optimization including the algorithm proposed for module processing, where the compiler serves to interpret graphs as well as to make a policy for effective execution of a neural network on NPUs. \emph{Neural network source code} represents a certain type of files including graph information of a neural network such as prototxt and TFLite formats. Through \emph{parse unit}, network parameters are extracted such as kernel size, stride, pad, layer type, feature-map size, module parameters and so on. With the extracted parameters, we can operate the proposed algorithm in \emph{optimization unit} which includes four phases: \emph{module detection, micro-instruction generation, branch reordering and branch processing (I/II)}. Here, \emph{module detection} is to detect the modules satisfying required conditions in a neural network, and \emph{micro-instruction generation} is to find the best sequence of micro-instructions per module defined in NPU. \emph{branch reordering} decides on a new order of branches to be processed in a module based on a proposed criteria. And \emph{branch processing I/II} decide an effective policy. Then, \emph{memory allocation unit} executes a resource allocation of on-chip memory based on the policy by the optimization unit. Finally, \emph{neural network object code} is generated as an output of \emph{object code generation unit}.

Since our main proposal is \emph{optimization unit} in Figure \ref{fig:004}, we explain it in detail. Algorithm \ref{alg:Optimization unit} shows a whole procedure of \emph{optimization unit} which has six steps such as module detection, operator sequence generation, branch reordering, MIFM (Module's Input Feature Map) memory size calculation, occupied MOFM (Module's Output Feature Map) memory size calculation, and branch processing. Here, a network graph $\Omega$ is used as an input in \emph{optimization unit}, and there are two types of outputs: the best sequence of micro-instructions, and the resource allocation information in on-chip memory per module. Each step of Algorithm \ref{alg:Optimization unit} is described in detail in Algorithm \ref{alg:module detection} through Algorithm \ref{alg:branch processing}.

First of all, the proposed algorithm has to detect all possible valid modules within a network. Different kinds of modules are designed for well-known neural network models, but we focus only on those modules which are suitable for efficient use of on-chip memory. Algorithm \ref{alg:module detection} is about how to detect modules that meet the required conditions as mentioned in the previous section. First, we skip modules configured by a long skip-connection since it is extremely difficult to manage them within on-chip memory efficiently. Then, the algorithm detects modules in a modified graph where all the long skip-connections are erased.

\begin{algorithm} []
   \caption{Optimization unit}
   \label{alg:Optimization unit}
\begin{algorithmic}
   \STATE {\bfseries Input:} Network graph ($\Omega$).
   \STATE {}

   \STATE \% [Step-1] module detection through $\Omega$
   \STATE \emph{ModuleDet($\Omega$)}
   \STATE {}

   \FOR{$m=1$ {\bfseries to} $len(modules)$}
   \STATE \% [Step-2] generate a sequence of micro-instructions for layers in the module
   \STATE \emph{Ops(m)=GenSeqOps(m)}
   \STATE {}

   \STATE \% [Step-3] branch re-ordering
   \STATE \emph{BrReordering(m)}
   \STATE {}

   \FOR{$b=1$ {\bfseries to} $len(branches~of~module)$}
   \FOR{$k=1$ {\bfseries to} $len(layers~of~branch)$}

   \STATE \% [Step-4] calculation of memory sizes of MIFM
   \STATE \emph{CalSizeMifmMem(k)}
   \ENDFOR

   \STATE {}
   \STATE \% [Step-5] calculation of the occupied memory size within MOFM according to $b$
   \STATE \emph{CalSizeMofmOccMem(b)}

   \STATE {}
   \STATE \% [Step-6] branch operation based on \emph{Ops}
   \STATE \emph{BrProcess}($b$, $I$) or \emph{BrProcess}($b$, $II$)
   \ENDFOR

   \STATE {}
   \ENDFOR

\end{algorithmic}
\end{algorithm}

\begin{algorithm} []
   \caption{Module detection, \emph{ModuleDet($\Omega$)}}
   \label{alg:module detection}
\begin{algorithmic}
   \STATE {\bfseries Input:} Network graph ($\Omega$) with a layer set ($\Psi$) and an edge set ($\Theta$).

   \FOR{$\nu$ in $\Theta$ }
   \IF {$\nu$ is not a \emph{long skip-connection}:}
   \STATE $\nu \in {\Theta}_e$.
   \ELSE
   \STATE ${\nu}_e$ = \emph{Disconnect}(${\nu}$)
   \STATE ${\nu}_e \in {\Theta}_e$, where ${\nu}_e$ is originated from off-chip memory.
   \ENDIF
   \ENDFOR

   \STATE {}
   \STATE {\bfseries Input:} Effective network graph (${\Omega}_e$) with a layer set ($\Psi$) and an effective edge set (${\Theta}_e$).

   \FOR{$\psi$ in $\Psi$ of ${\Omega}_e$}
   \IF {$\psi$ == \emph{merge layer}:}
   \STATE $\sigma$ = \emph{SearchStartLayer}($\psi$)
   \STATE $\omega$ = \emph{ExtractModuleParam}($\sigma$, $\psi$)
   \ENDIF
   \ENDFOR

\end{algorithmic}
\end{algorithm}

\begin{figure}[t]
\begin{center}
\includegraphics[width=0.9\linewidth]{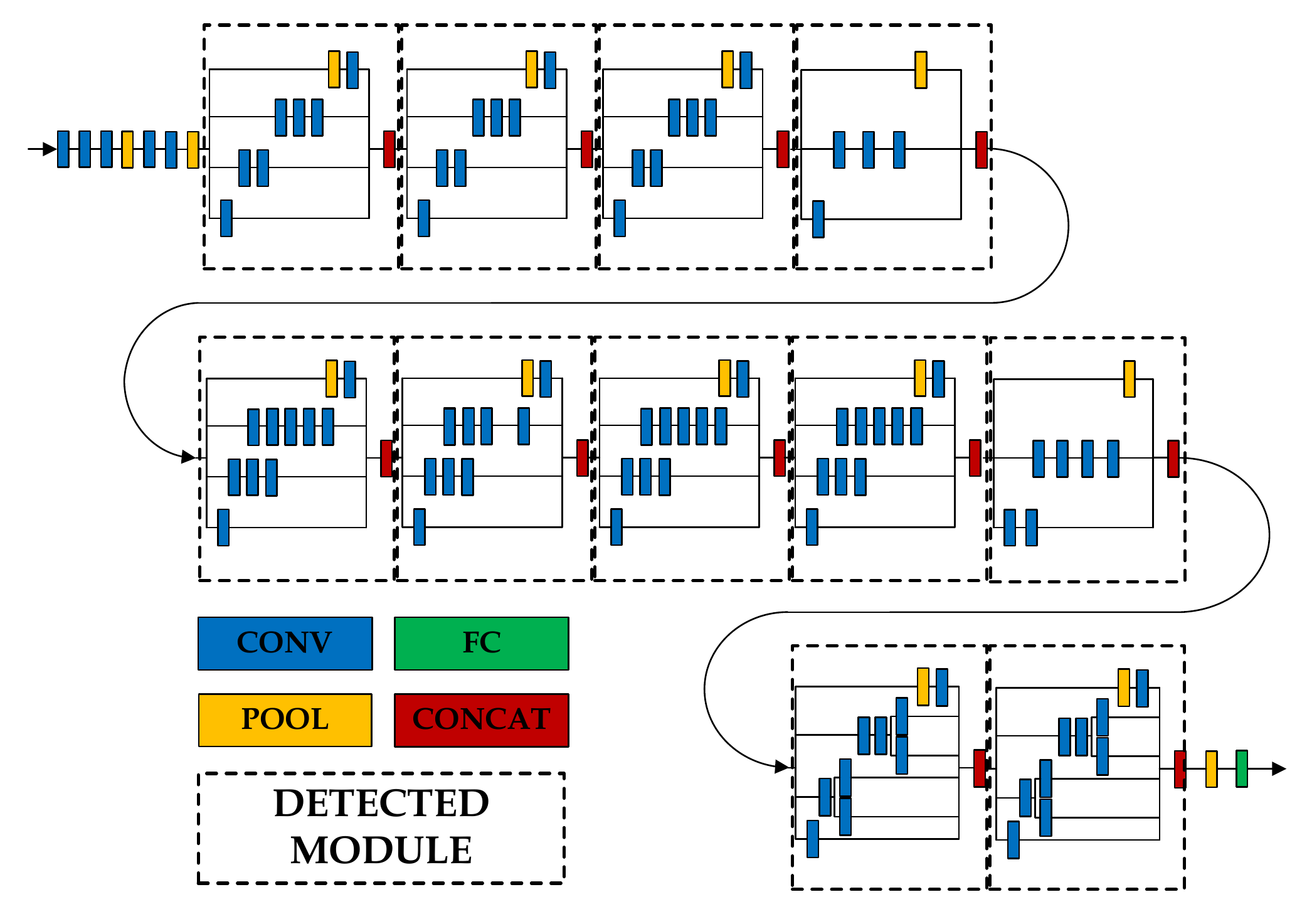}
\end{center}
   \caption{Illustration of results of the module detection algorithm in Algorithm \ref{alg:module detection}.}
\label{fig:005}
\end{figure}

\begin{figure}[h]
\begin{center}
\includegraphics[width=0.8\linewidth]{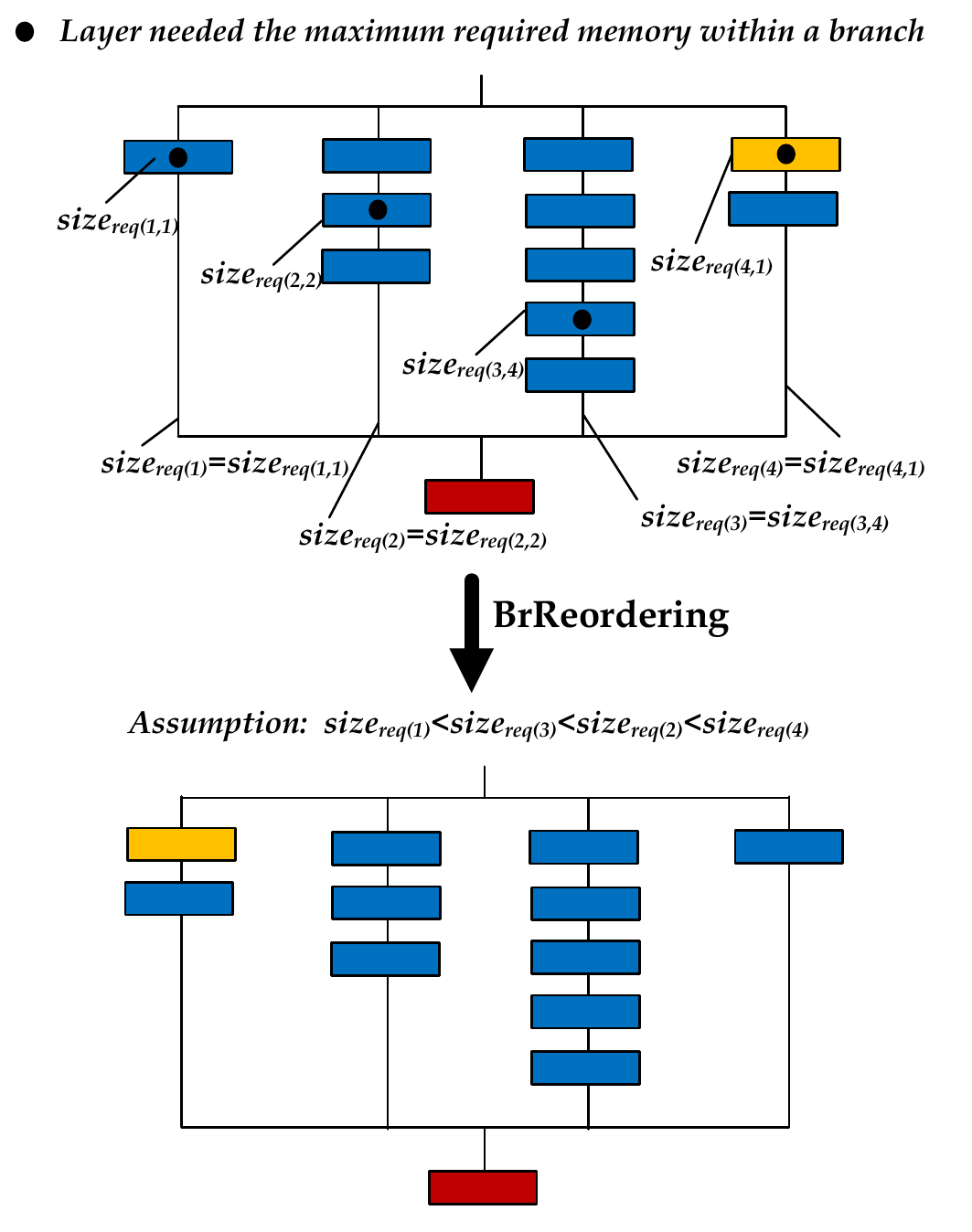}
\end{center}
   \caption{Illustration of the branch re-ordering algorithm in Algorithm \ref{alg:br reorder}}
\label{fig:006}
\end{figure}

Figure \ref{fig:005} shows the results of the module detection algorithm being applied to Inception-V3. Here we can see that 11 modules are detected by the Algorithm \ref{alg:module detection}. After detecting modules, the optimization unit generates a sequence of micro-instructions for whole layers of each module. Since this step is very hardware-dependent, it is hard to explain details in this paper. In the optimization unit, anyway, the best sequence of micro-instructions should be found in order to make the layers of each module processed efficiently in a specific hardware.

\begin{algorithm} [b]
   \caption{Re-ordering of branches, \emph{BrReordering(m)}}
   \label{alg:br reorder}
\begin{algorithmic}
   \STATE {\bfseries Input:} module parameters $\omega$.

   \STATE {}
   \FOR{$b=1$ {\bfseries to} $len(branches~of~module)$}
   \FOR{$k=1$ {\bfseries to} $len(layers~of~branch)$}
   \STATE \% calculating a size of the required memory except MIFM and MOFM for each layer
   \STATE $size_{req(b,k)} = CalcSizeReqMem(b,k)$
   \ENDFOR
   \STATE $size_{req(b)}~=~max(size_{req(b,:)})$
   \ENDFOR
   \STATE \% Sorting the indexes of branches in descending order
   \STATE \emph{Sort}($size_{req(:)}$)

\end{algorithmic}
\end{algorithm}

As the $3^{rd}$ step, we execute a re-ordering process of branches for all modules through Algorithm \ref{alg:br reorder}. Firstly, we calculate a required memory size of each branch within a module. Here the required memory size of a layer is calculated by accumulating sizes of IFM, OFM, internal working memories, and weights in a function of $CalcSizeReqMem$, but except MIFM and MOFM. And the required memory size of a branch is determined by the largest of the memory sizes required for layers on the branch. Finally, we change the processing order of the branches in descending order of the memory size required by each branch. Figure \ref{fig:006} shows an example including a module with four branches. When it is assumed $size_{req(1)}<size_{req(3)}<size_{req(2)}<size_{req(4)}$, \emph{BrReordering} changes the order of branches as follows: $Br(4)$, $Br(2)$, $Br(3)$, and $Br(1)$ in the right plot of Figure \ref{fig:006}. Through \emph{BrReordering}, we can allocate more available resource in on-chip memory for the branch which requires a larger size of memory.

\begin{algorithm} [b]
   \caption{Calculation of MIFM size for the $k^{th}$ layer, \emph{CalSizeMifmMem(k)}}
   \label{alg:size calculation}
\begin{algorithmic}
   \STATE {\bfseries Input:} module parameters $\omega$, effective depth of MIFM for the $k^{th}$ layer in module $d_i(k)$.

   \STATE {}
   \FOR {$i=1$ {\bfseries to} $d_i(k)$}
   \STATE \% calculating MIFM size for the module with $\omega$
   \STATE $size_{mifm(k,i)} = calcMifmMemSize(k,i)$
   \STATE $offset_{mifm(k,i)} = calcMifmMemOffset(k,i)$
   \ENDFOR
   \STATE

\end{algorithmic}
\end{algorithm}

\begin{figure}[t]
\begin{center}
\includegraphics[width=0.8\linewidth]{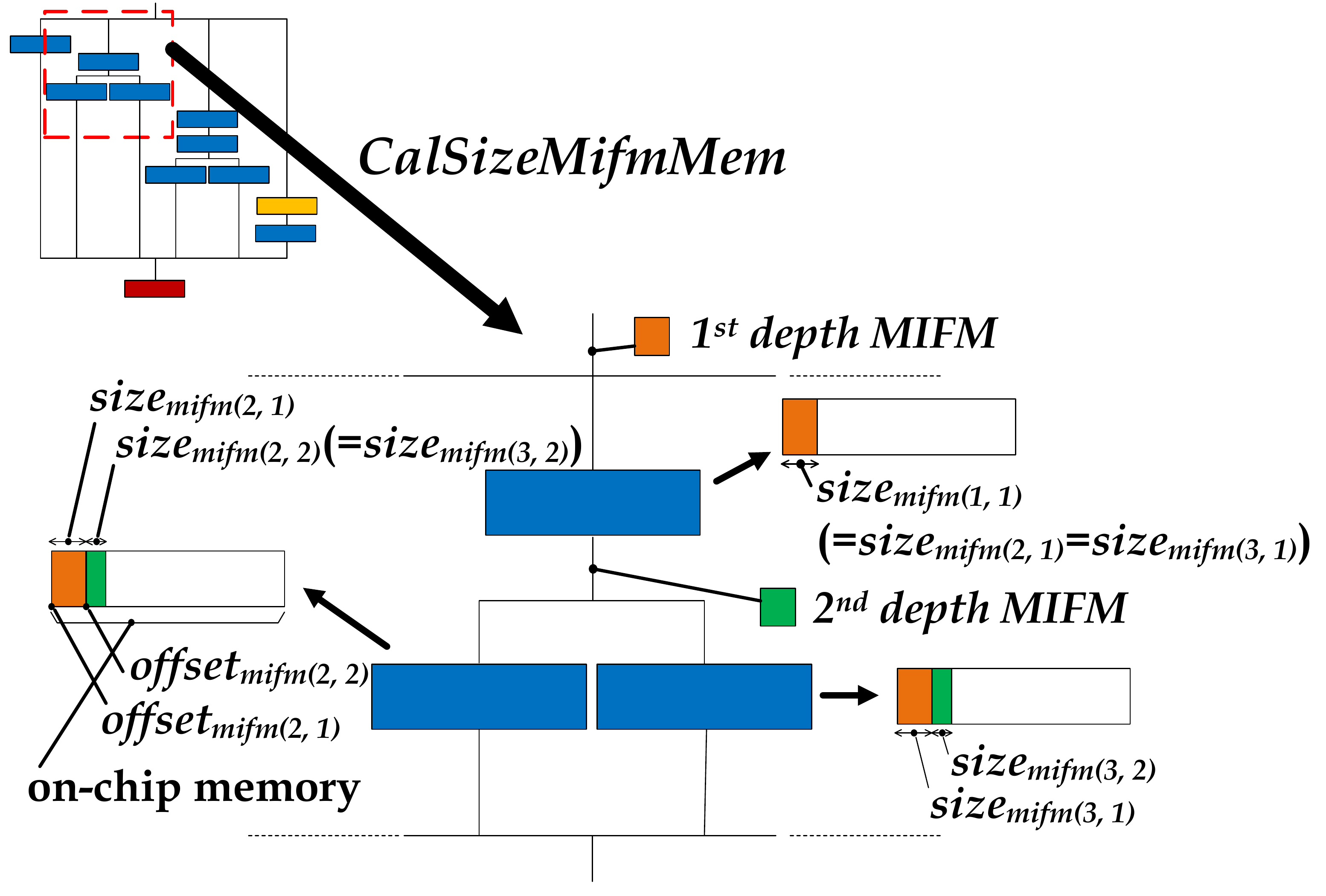}
\end{center}
   \caption{Illustration of the MIFM calculation algorithm in Algorithm \ref{alg:size calculation}}
\label{fig:007}
\end{figure}

At the next step, we need to calculate the size of MIFM and to allocate an offset in order to share a memory region without any collision during module processing. As shown in Algorithm \ref{alg:size calculation}, sizes and offsets are calculated at the $k^{th}$ layer for all possible depths, $d_i(k)$, in the module. In Figure \ref{fig:007}, we look at the red-colored box as an example of MIFM calculation algorithm, where it is the second branch with three layers in the module. If you look at the $1^{st}$ layer, only $size_{mifm(1,1)}$ is considered because it is included in the only $1^{st}$ depth module. Because $size_{mifm(1,1)}$ is exactly same with the size of the previous branch, of course, the memory region is directly handed over from the previous branch. For the $2^{nd}$ and the $3^{rd}$ layers within the $2^{nd}$ depth module, we need to consider the second shared memory region of $size_{mifm(2,2)}$ and $offset_{mifm(2,2)}$. After operating the $3^{rd}$ layer, we can release the second shared memory region of $size_{mifm(2,2)}$ and $offset_{mifm(2,2)}$ because there is no need in the next branches.

\begin{algorithm} [b]
   \caption{Calculation of the occupied memory size of MOFM at the $b^{th}$ branch, \emph{CalSizeMofmOccMem(b)}}
   \label{alg:MofmOcc calculation}
\begin{algorithmic}
   \STATE {\bfseries Input:} module parameters $\omega$.

   \STATE {}
   \STATE {$size_{occu(b)}$ = 0}
   \FOR {$i=1$ {\bfseries to} $b-1$}
   \STATE $size_{occu(b)} += calcMofmMemSize(i)$
   \STATE $offset_{occu(b)} = calcMofmMemOffset(i)$
   \ENDFOR
   \STATE

\end{algorithmic}
\end{algorithm}

\begin{figure}[]
\begin{center}
\includegraphics[width=0.7\linewidth]{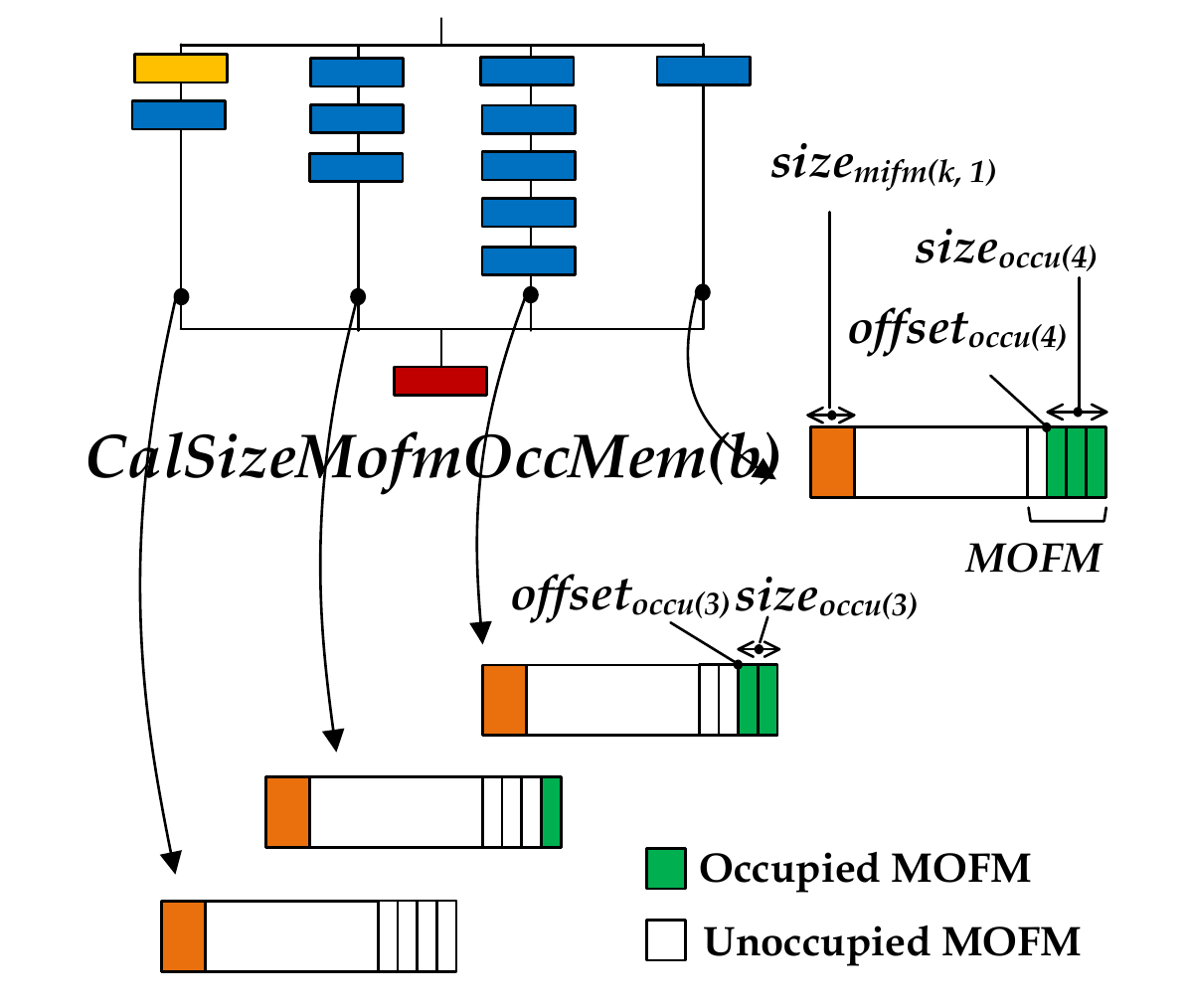}
\end{center}
   \caption{Illustration of the occupied MOFM calculation algorithm in Algorithm \ref{alg:MofmOcc calculation}}
\label{fig:008}
\end{figure}

As the $5^{th}$ step, we calculate the occupied memory sizes within MOFM at the $b^{th}$ branch in Algorithm \ref{alg:MofmOcc calculation}. It can be simply calculated by accumulating the output size of the last layer on all previous branches. It means $size_{occu(b)}$ is the size of occupied region of the MOFM for the current branch as shown in Figure \ref{fig:008}. It is important to exactly calculate the size of the occupied region because the region (greed colored region in Figure \ref{fig:008}) in on-chip memory cannot be utilized for the current branch operations.

\begin{figure}[]
\begin{center}
\includegraphics[width=0.8\linewidth]{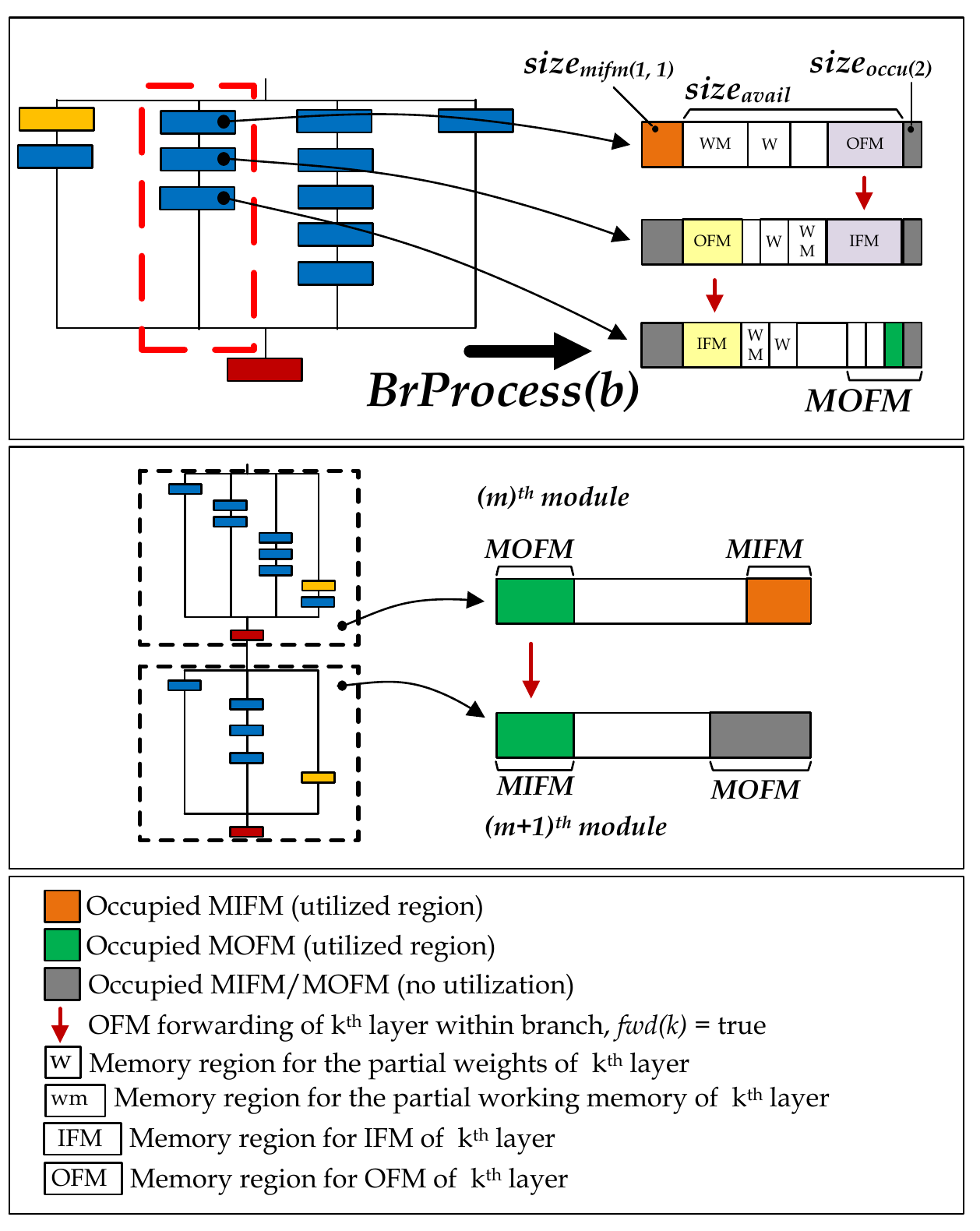}
\end{center}
   \caption{Illustration of the branch processing algorithm in Algorithm \ref{alg:branch processing}}
\label{fig:009}
\end{figure}

\begin{algorithm} [t]
   \caption{Branch processing, \emph{BrProcess}($b$, $opt$)}
   \label{alg:branch processing}
\begin{algorithmic}
   \STATE {\bfseries Input:} Branch index $b$, and MIFM depth $d_i(k)$, and pre-occupied MOFM size $size_{occu(b)}$
   \STATE {}

   \STATE  $fwd_{(0)}$=false;
   \STATE \% estimating the forwarding status of OFM for the $k^{th}$ layer on the branch
   \FOR {$k=1$ {\bfseries to} $len(layers~of~branch)$}
   \STATE \% calculating the available memory size
   \IF {$opt$ = $'II'$}
   \STATE $size_{occu(b)}$=0
   \ENDIF
   \STATE {$size_{avail(b, k)}$ = $size_{mem}$} - $\sum_{m=1}^{d_i(k)} size_{mifm(k, m)}$ - $size_{occu(b)}$
   \IF {($fwd_{(k-1)}$):}
   \STATE {$size_{req(b, k)}$ = $size_{IFM.Full}$ + $size_{OFM.Full}$ + $size_{WM.Partial}$ + $size_{W.Partial}$ }

   \IF {$size_{req(b, k)} \leq size_{avail(b, k)}$:}
   \STATE {$fwd_{(k)}$ = true;}
   \ELSE
   \STATE {$fwd_{(k)}$ = false;}
   \ENDIF

   \ELSE
   \STATE {$size_{req(b, k)}$ = $size_{IFM.Partial}$ + $size_{OFM.Full}$ + $size_{WM.Partial}$ + $size_{W.Partial}$ }

   \IF {$size_{req(b, k)} \leq size_{avail(b, k)}$:}
   \STATE {$fwd_{(k)}$ = true;}
   \ELSE
   \STATE {$fwd_{(k)}$ = false;}
   \ENDIF
   \ENDIF

   \IF {($opt$ = $'II'$) and ($k$==$len(layers~of~branch)$)}
   \STATE {$fwd_{(k)}$ = false;}
   \STATE {\% $size_{req(b, k)}$ can be calculated with $size_{OFM.Partial}$, not $size_{OFM.Full}$. }
   \ENDIF

   \ENDFOR

\end{algorithmic}
\end{algorithm}

The branch processing is conducted at the last step as shown in Algorithm \ref{alg:branch processing}. We propose two types of branch processing: (I) default algorithm considering both MIFM and MOFM, and (II) optional algorithm considering only MIFM. The optimization unit adaptively chooses one between (I) and (II) according to the required memory size of a module. The branch processing (I) can support an operation where MOFM of the former module is directly forwarded to MIFM of the latter module (= sharing memory between consecutive modules), but we have to use less memory region ($size_{avail}$ in Figure \ref{fig:009}) within branch operation because it needs the occupied memory region for both MIFM and MOFM in the module. In other hands, the branch processing (II) is an optional algorithm applicable when a required memory size $size_{req(b)}$ for the $b^{th}$ branch is larger than that of available on-chip memory $size_{avail(b,k)}$. It is because we can use additional memory region of $size_{occu(b)}$ for a branch processing and because we can also get more available memory by considering $size_{OFM.Partial}$, not $size_{OFM.Full}$ at the last layer on the branch. That is, we give up the shared memory region of MOFM in order to increase $size_{avail}$, instead we get a benefit only from sharing MIFM on the branch. More detail operations are explained in Algorithm \ref{alg:branch processing}, where $fwd(k)$ denotes a flag of an OFM forwarding of the $k^{th}$ layer. Figure \ref{fig:009} shows the example of the branch processing (I), where we consider the second branch operation including three layers within a module. At the $1^{st}$ layer, $size_{mifm(1,1)}$ means the shared memory for MIFM in the module, and $size_{occu(2)}$ is mapped to the occupied region by OFM of last layer in the $1^{st}$ branch. That is, we can use only $size_{avail}$ which is defined in Figure \ref{fig:009} for the $1^{st}$ layer. If the total size including OFM, working memory (WM) and weights (W) is equal to or less than $size_{avail}$, OFM can be directly forwarded to IFM of the $2^{nd}$ layer, as shown in Figure \ref{fig:009}. At the $3^{rd}$ layer by the same sequences, IFM is a heritage region from OFM of the second layer and OFM is stored as the green colored region within MOFM region. By using the branch processing (I), therefore, we can completely erase off-chip memory accesses during the processing within a module. The algorithm also provides the removal of the off-chip accesses between consecutive modules by forwarding the MOFM of the former module to the MIFM of the latter module in neural networks, as shown in the second plot of Figure \ref{fig:009}.


\begin{figure*}[t]
\begin{center}
\includegraphics[width=0.8\linewidth]{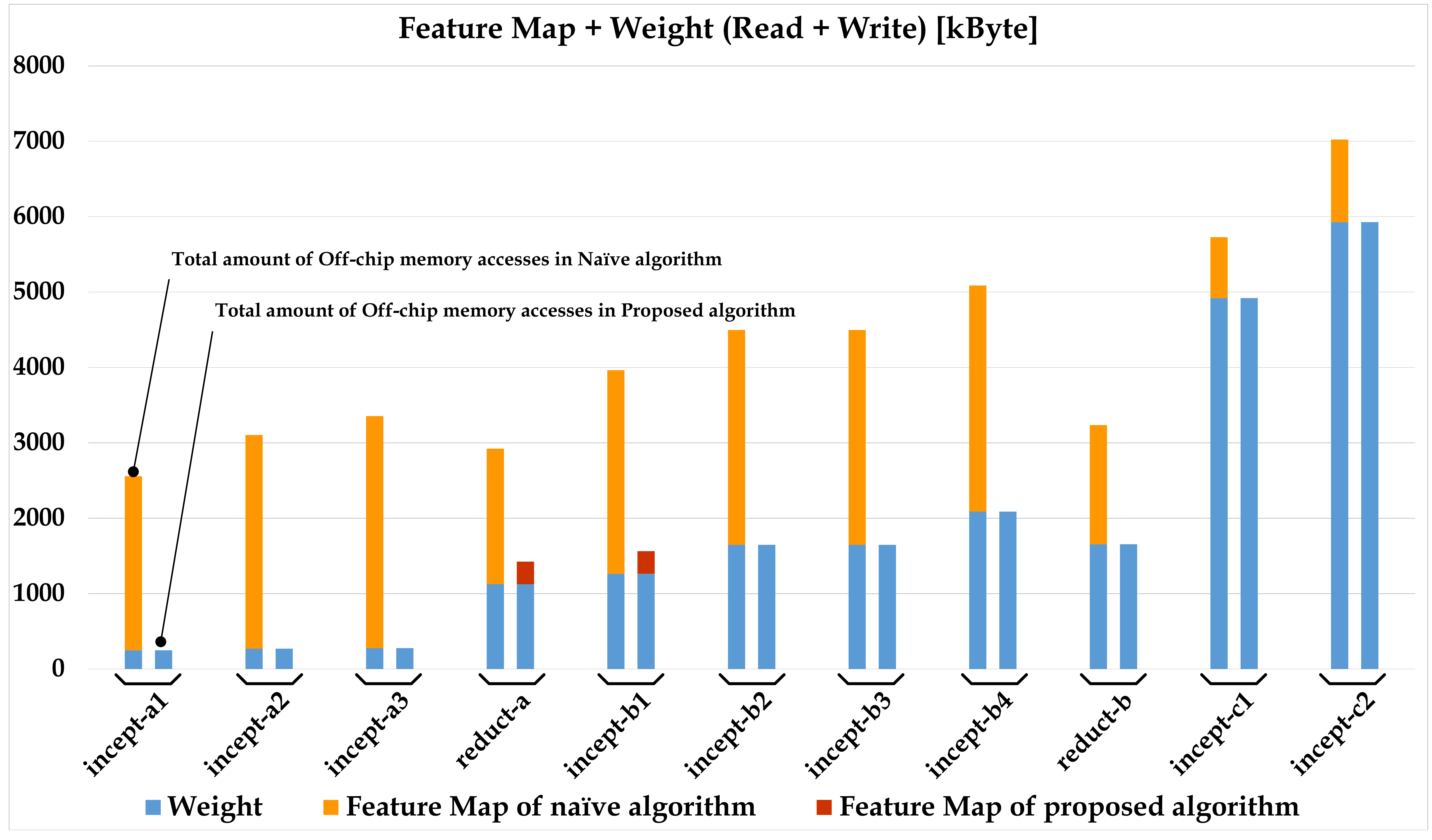}
\end{center}
   \caption{Comparison of total data sizes accessed the off-chip memory for Naive and Proposed algorithms, where total data size was summed for both weights and feature-maps.}
\label{fig:010}
\end{figure*}

\section{Evaluation}
\label{sec:evaluation}

To evaluate the proposed algorithm, a representative CNN model, Inception-V3 \cite{szegedy2016rethinking} with an input image of $299 \times 299$, has been selected. The target neural processor and its features are as follows: 1) Samsung’s NPU \cite{song2018makalu} in Exynos has 1024 multiply/accumulate (MAC) units on 16 MAAs (multiply/accumulate arrays), and on-chip memory of 1,024 Kbyte which contains IFMs, OFMs, weights and temporary WMs. The NPU also has 3 parallelism: First, IFMs are divided and fetched into four chunks along channel. Second, OFMs in the form of 4x4 patch in a MAA are computed in parallel. Lastly, a weight kernel is copied to 16 kernels for parallel operation on 16 MAAs. 2) It is not essential that all weight kernels of a layer have to be in on-chip memory. The NPU can make 16 OFM channels in parallel if partial weights for 16 MAAs are in on-chip memory. The partial weights can be read and written in a double buffering manner to effectively hide memory access time. 3) We assume that all weights and feature-maps are 8-bit quantized in the evaluation, even though the NPU supports other precisions.

\begin{figure}[]
\begin{center}
\includegraphics[width=1.0\linewidth]{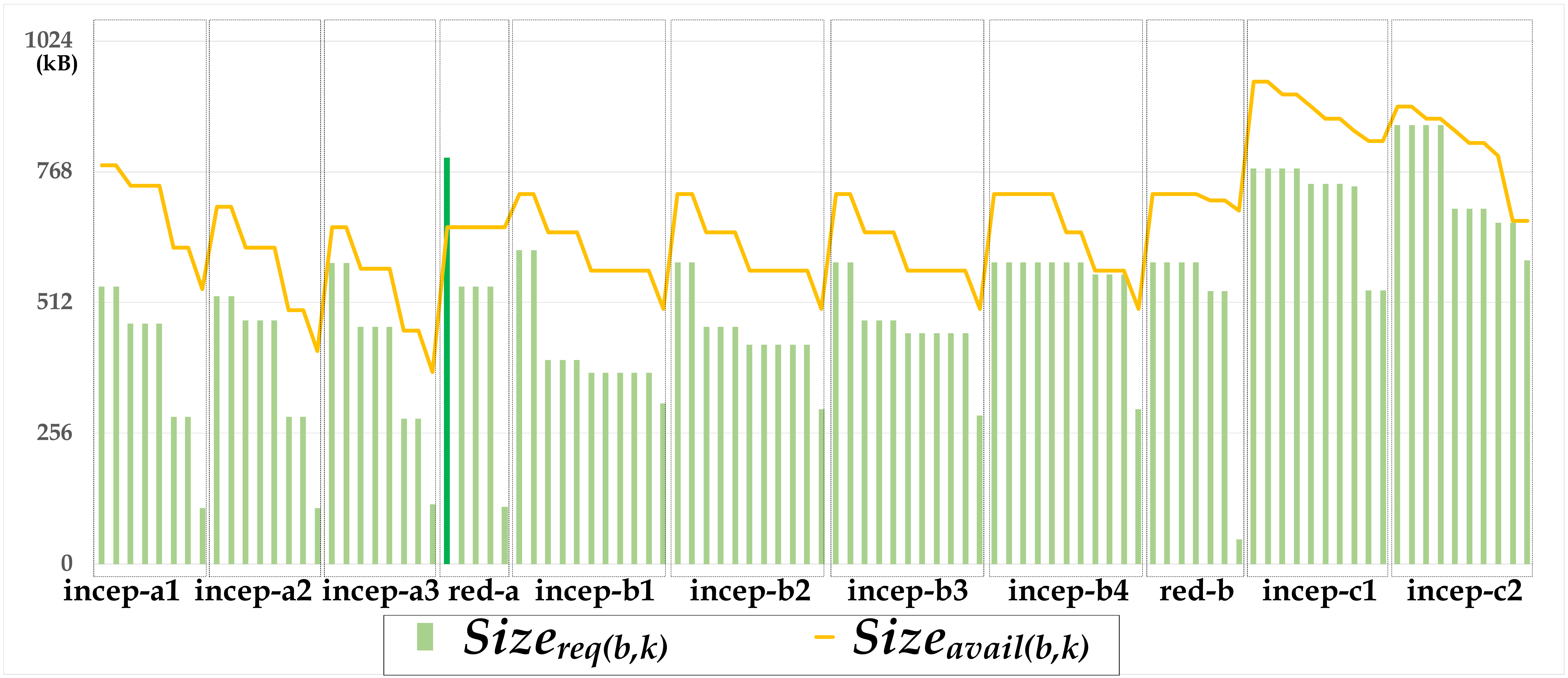}
\end{center}
   \caption{Illustration of the relationship between $size_{req(b,k)}$ and $size_{avail(b,k)}$ through the proposed algorithm at the $k^{th}$ layer on the $b^{th}$ branch for 11 modules in Inception-V3.}
\label{fig:011}
\end{figure}

\begin{table*} [t]
\begin{center}
\begin{tabular}{ c|c|c|c|c|c|c|c|c|c|c|c }
  \hline
  \multirow{3}{*}{}& \multicolumn{7}{c|}{Amount of } & \multicolumn{4}{c}{(For FM) Number of} \\
  & \multicolumn{7}{c|}{off-chip memory accesses [KB]} & \multicolumn{4}{c}{off-chip memory accesses } \\
  \cline{2-12}
  & \multirow{2}{*}{W} & \multicolumn{2}{c|}{Naive} & \multicolumn{2}{c|}{Proposed} & {Overall} & {FM} & \multicolumn{2}{c|}{Naive}& \multicolumn{2}{c}{Proposed} \\
  \cline{3-6} \cline{9-12}
  & & FM & Overall & FM & Overall & {Ratio [\%]} & {Ratio [\%]} & Read & Write & Read & Write \\
  \hline
  inception-a1&  249  &  2308.5  &  2557.5  &  0  &  249  &  9.74 &0.00 &8	&8	&0	&0\\
  \hline
  inception-a2&  270  &  2835  &  3105  &  0  &  270  &  8.70 &0.00  &8	&8	&0	&0\\
  \hline
  inception-a3&  277.5  &  3078  &  3355.5  &  0  &  277.5  &  8.27 &0.00 &8	&8	&0	&0\\
  \hline
  reduction-a&  1125  &  1798.5  &  2923.5  &  300  &  1425  &  48.74 &16.68 &5	&5	&0	&3 \\
  \hline
  inception-b1&  1264  &  2700  &  3964  &  300  &  1564  &  39.46  &11.11 &11	&11	&1	&0\\
  \hline
  inception-b2&  1648  &  2850  &  4498  &  0  &  1648  &  36.64  &0.00 &11	&11	&0	&0 \\
  \hline
  inception-b3&  1648  &  2850  &  4498  &  0  &  1648  &  36.64  &0.00 &11	&11	&0	&0 \\
  \hline
  inception-b4&  2088  &  3000  &  5088  &  0  &  2088  &  41.04  &0.00 &11	&11	&0	&0 \\
  \hline
  reduction-b&  1656  &  1580  &  3236  &  0  &  1656  &  51.17  &0.00 &7	&7	&0	&0\\
  \hline
  inception-c1&  4920  &  808  &  5728  &  0  &  4920  &  85.89  &0.00 &10	&10	&0	&0 \\
  \hline
  inception-c2&  5928  &  1096  &  7024  &  0  &  5928  &  84.40 &0.00 &10	&10	&0	&0 \\
  \hline
  total&  21073.5  &  24904  &  45977.5  &  600  &  21673.5  &  47.14 &2.41 &100	&100	&1	&3 \\
  \hline

\end{tabular}
\end{center}
\caption{Summary about the amount and the number of the off-chip accesses for 11 modules in Inception-V3, where FM and W mean a feature-map and a weight, respectively.}
\label{tbl:001}
\end{table*}

Figure \ref{fig:010} shows the total amount of data with off-chip memory access for two algorithms: Naive and Proposed algorithms. Here, it is assumed that Naive algorithm has to access off-chip memory to read and write feature-maps for every layers. The results show that the amount of feature-map with off-chip memory access is almost gone and the amount of weight with off-chip memory access remains the same. Therefore, the proposed algorithm reduces the total amount of data with off-chip memory access by almost half. By applying the proposed algorithm, only \emph{reduction-a} and \emph{Inception-b1} among 11 modules in Inception-V3 have a little amount of feature-map with off-chip memory access. The reason can be explained as follows: In \emph{reduction-a} module, we operated $BrProcessing(b, II)$ which needs to access off-chip memory three times at the end of branches because $size_{req}$ was not satisfied with $size_{avail}$ for the module, as shown in Figure \ref{fig:011}. That is why the off-chip accesses in \emph{reduction-a} happen. And then, there exists one time of off-chip memory access at the start of the \emph{Inception-b1} module according to the result of the former module, \emph{reduction-a}. Figure \ref{fig:011} shows the important information related to $size_{req(b,k)}$ and $size_{avail(b,k)}$ in our proposed algorithm. Firstly, branches in each module are re-ordered by $size_{req(b,k)}$. And $size_{avail(b,k)}$ also decreases as the index of branch in the module increases because $size_{occu(b)}$ increases. Moreover, we can see the relationship between $size_{req(b,k)}$ and $size_{avail(b,k)}$ used for the criteria to select $BrProcessing(b, I)$ or $BrProcessing(b, II)$.

Consequently, Table \ref{tbl:001} summarizes the overall amount of off-chip accesses for every modules in Inception-V3, and an reduction ratio is calculated to 47.14 \% (=21673.5 / 45977.5). And The table also represents the amount and the number of off-chip accesses for only a feature-map in Inception-V3. We can know that the proposed algorithm can achieve a great reduction up to 97.59 \% in terms of the amount, and can effectively reduce to 1/50 (= 4/200) in terms of the number of accesses for all modules.

\section{Conclusion}
\label{sec:conclusion}

In the paper, we proposed a simple method for energy-efficient and real-time processing of NPUs through the reduction of off-chip memory accesses. To achieve it, we focused on the modules used for convolutional neural networks that have multiple branches and a merge layer. In the algorithm, the key ideas consisted of module detection ignoring long skip-connections, branch re-ordering for utilizing available memory maximally, assignment of MIFM and MOFM to share between modules, and branch processing. For Inception-V3 on Samsung's NPU in Exynos, we showed the proposed algorithm achieved 97.59 \% reduction in the amount of data to require off-chip access, and reduced the number of off-chip accesses by 1/50. Finally, we think the proposed algorithm can be a powerful solution to increase the efficiency of NPUs when processing various convolutional neural networks.


\nocite{langley00}

\bibliography{icml2019-compiler}

\begin{thebibliography}{25}
\providecommand{\natexlab}[1]{#1}
\providecommand{\url}[1]{\texttt{#1}}
\expandafter\ifx\csname urlstyle\endcsname\relax
  \providecommand{\doi}[1]{doi: #1}\else
  \providecommand{\doi}{doi: \begingroup \urlstyle{rm}\Url}\fi

\bibitem[ARM-ML-processor()]{Marvin2018Online}
ARM-ML-processor.
\newblock industry-leading performance and efficiency for inference at the
  edge.
\newblock
  \url{https://developer.arm.com/products/processors/machine-learning/arm-ml-processor}.

\bibitem[Courbariaux et~al.(2014)Courbariaux, Bengio, and
  David]{courbariaux2014training}
Courbariaux, M., Bengio, Y., and David, J.-P.
\newblock Training deep neural networks with low precision multiplications.
\newblock \emph{arXiv preprint arXiv:1412.7024}, 2014.

\bibitem[Courbariaux et~al.(2015)Courbariaux, Bengio, and
  David]{courbariaux2015binaryconnect}
Courbariaux, M., Bengio, Y., and David, J.-P.
\newblock Binaryconnect: Training deep neural networks with binary weights
  during propagations.
\newblock In \emph{Advances in Neural Information Processing Systems}, pp.\
  3123--3131, 2015.

\bibitem[Gupta et~al.(2015)Gupta, Agrawal, Gopalakrishnan, and
  Narayanan]{gupta2015deep}
Gupta, S., Agrawal, A., Gopalakrishnan, K., and Narayanan, P.
\newblock Deep learning with limited numerical precision.
\newblock In \emph{Proceedings of the 32nd International Conference on Machine
  Learning (ICML-15)}, pp.\  1737--1746, 2015.

\bibitem[Gysel et~al.(2016)Gysel, Motamedi, and Ghiasi]{gysel2016hardware}
Gysel, P., Motamedi, M., and Ghiasi, S.
\newblock Hardware-oriented approximation of convolutional neural networks.
\newblock \emph{arXiv preprint arXiv:1604.03168}, 2016.

\bibitem[Han et~al.(2015)Han, Mao, and Dally]{han2015deep}
Han, S., Mao, H., and Dally, W.~J.
\newblock Deep compression: Compressing deep neural networks with pruning,
  trained quantization and huffman coding.
\newblock \emph{arXiv preprint arXiv:1510.00149}, 2015.

\bibitem[Han et~al.(2016{\natexlab{a}})Han, Liu, Mao, Pu, Pedram, Horowitz, and
  Dally]{han2016eie}
Han, S., Liu, X., Mao, H., Pu, J., Pedram, A., Horowitz, M.~A., and Dally,
  W.~J.
\newblock Eie: efficient inference engine on compressed deep neural network.
\newblock In \emph{Computer Architecture (ISCA), 2016 ACM/IEEE 43rd Annual
  International Symposium on}, pp.\  243--254. IEEE, 2016{\natexlab{a}}.

\bibitem[Han et~al.(2016{\natexlab{b}})Han, Pool, Narang, Mao, Tang, Elsen,
  Catanzaro, Tran, and Dally]{han2016dsd}
Han, S., Pool, J., Narang, S., Mao, H., Tang, S., Elsen, E., Catanzaro, B.,
  Tran, J., and Dally, W.~J.
\newblock Dsd: Regularizing deep neural networks with dense-sparse-dense
  training flow.
\newblock \emph{arXiv preprint arXiv:1607.04381}, 2016{\natexlab{b}}.

\bibitem[He et~al.(2016)He, Zhang, Ren, and Sun]{he2016deep}
He, K., Zhang, X., Ren, S., and Sun, J.
\newblock Deep residual learning for image recognition.
\newblock In \emph{Proceedings of the IEEE conference on computer vision and
  pattern recognition}, pp.\  770--778, 2016.

\bibitem[Howard et~al.(2017)Howard, Zhu, Chen, Kalenichenko, Wang, Weyand,
  Andreetto, and Adam]{howard2017mobilenets}
Howard, A.~G., Zhu, M., Chen, B., Kalenichenko, D., Wang, W., Weyand, T.,
  Andreetto, M., and Adam, H.
\newblock Mobilenets: Efficient convolutional neural networks for mobile vision
  applications.
\newblock \emph{arXiv preprint arXiv:1704.04861}, 2017.

\bibitem[Hubara et~al.(2016)Hubara, Courbariaux, Soudry, El-Yaniv, and
  Bengio]{hubara2016quantized}
Hubara, I., Courbariaux, M., Soudry, D., El-Yaniv, R., and Bengio, Y.
\newblock Quantized neural networks: Training neural networks with low
  precision weights and activations.
\newblock \emph{arXiv preprint arXiv:1609.07061}, 2016.

\bibitem[Iandola et~al.(2016)Iandola, Han, Moskewicz, Ashraf, Dally, and
  Keutzer]{iandola2016squeezenet}
Iandola, F.~N., Han, S., Moskewicz, M.~W., Ashraf, K., Dally, W.~J., and
  Keutzer, K.
\newblock Squeezenet: Alexnet-level accuracy with 50x fewer parameters and< 0.5
  mb model size.
\newblock \emph{arXiv preprint arXiv:1602.07360}, 2016.

\bibitem[Judd et~al.(2015)Judd, Albericio, Hetherington, Aamodt, Jerger,
  Urtasun, and Moshovos]{judd2015reduced}
Judd, P., Albericio, J., Hetherington, T., Aamodt, T., Jerger, N.~E., Urtasun,
  R., and Moshovos, A.
\newblock Reduced-precision strategies for bounded memory in deep neural nets.
\newblock \emph{arXiv preprint arXiv:1511.05236}, 2015.

\bibitem[Kim et~al.(2018)Kim, Yim, Ha, Lee, and Kang]{kim2018convolutional}
Kim, D., Yim, H.~Y., Ha, S., Lee, C., and Kang, I.
\newblock Convolutional neural network quantization using generalized gamma
  distribution.
\newblock \emph{arXiv preprint arXiv:1810.13329}, 2018.

\bibitem[Lin et~al.(2016)Lin, Talathi, and Annapureddy]{lin2016fixed}
Lin, D., Talathi, S., and Annapureddy, S.
\newblock Fixed point quantization of deep convolutional networks.
\newblock In \emph{International Conference on Machine Learning}, pp.\
  2849--2858, 2016.

\bibitem[Lin et~al.(2013)Lin, Chen, and Yan]{lin2013network}
Lin, M., Chen, Q., and Yan, S.
\newblock Network in network.
\newblock \emph{arXiv preprint arXiv:1312.4400}, 2013.

\bibitem[Long et~al.(2015)Long, Shelhamer, and Darrell]{long2015fully}
Long, J., Shelhamer, E., and Darrell, T.
\newblock Fully convolutional networks for semantic segmentation.
\newblock In \emph{Proceedings of the IEEE Conference on Computer Vision and
  Pattern Recognition}, pp.\  3431--3440, 2015.

\bibitem[Ma et~al.(2018)Ma, Zhang, Zheng, and Sun]{ma2018shufflenet}
Ma, N., Zhang, X., Zheng, H.-T., and Sun, J.
\newblock Shufflenet v2: Practical guidelines for efficient cnn architecture
  design.
\newblock \emph{arXiv preprint arXiv:1807.11164}, 2018.

\bibitem[Ren et~al.(2015)Ren, He, Girshick, and Sun]{ren2015faster}
Ren, S., He, K., Girshick, R., and Sun, J.
\newblock Faster r-cnn: Towards real-time object detection with region proposal
  networks.
\newblock In \emph{Advances in neural information processing systems}, pp.\
  91--99, 2015.

\bibitem[Sandler et~al.(2018)Sandler, Howard, Zhu, Zhmoginov, and
  Chen]{sandler2018mobilenetv2}
Sandler, M., Howard, A., Zhu, M., Zhmoginov, A., and Chen, L.-C.
\newblock Mobilenetv2: Inverted residuals and linear bottlenecks.
\newblock In \emph{Proceedings of the IEEE Conference on Computer Vision and
  Pattern Recognition}, pp.\  4510--4520, 2018.

\bibitem[Song et~al.()Song, Cho, Park, Jang, Lee, Song, Lee, and
  Kang]{song2018makalu}
Song, J., Cho, Y., Park, J.-S., Jang, J.-W., Lee, S., Song, J.-H., Lee, J.-G.,
  and Kang, I.
\newblock An 11.5tops/w 1024-mac butterfly structure dual-core sparsity-aware
  neural processing unit in 8nm flagship mobile soc.
\newblock In \emph{accepted for 2019 IEEE International Solid-State Circuits
  Conference (ISSCC)}.

\bibitem[Szegedy et~al.(2015)Szegedy, Liu, Jia, Sermanet, Reed, Anguelov,
  Erhan, Vanhoucke, and Rabinovich]{szegedy2015going}
Szegedy, C., Liu, W., Jia, Y., Sermanet, P., Reed, S., Anguelov, D., Erhan, D.,
  Vanhoucke, V., and Rabinovich, A.
\newblock Going deeper with convolutions.
\newblock In \emph{Proceedings of the IEEE conference on computer vision and
  pattern recognition}, pp.\  1--9, 2015.

\bibitem[Szegedy et~al.(2016)Szegedy, Vanhoucke, Ioffe, Shlens, and
  Wojna]{szegedy2016rethinking}
Szegedy, C., Vanhoucke, V., Ioffe, S., Shlens, J., and Wojna, Z.
\newblock Rethinking the inception architecture for computer vision.
\newblock In \emph{Proceedings of the IEEE conference on computer vision and
  pattern recognition}, pp.\  2818--2826, 2016.

\bibitem[Tan et~al.(2018)Tan, Chen, Pang, Vasudevan, and Le]{tan2018mnasnet}
Tan, M., Chen, B., Pang, R., Vasudevan, V., and Le, Q.~V.
\newblock Mnasnet: Platform-aware neural architecture search for mobile.
\newblock \emph{arXiv preprint arXiv:1807.11626}, 2018.

\bibitem[Zhang et~al.(2016)Zhang, Du, Zhang, Lan, Liu, Li, Guo, Chen, and
  Chen]{zhang2016cambricon}
Zhang, S., Du, Z., Zhang, L., Lan, H., Liu, S., Li, L., Guo, Q., Chen, T., and
  Chen, Y.
\newblock Cambricon-x: An accelerator for sparse neural networks.
\newblock In \emph{The 49th Annual IEEE/ACM International Symposium on
  Microarchitecture}, pp.\ ~20. IEEE Press, 2016.

\end{thebibliography}
\bibliographystyle{icml2019}

\end{document}